\documentclass[10pt,twocolumn,letterpaper]{article}

\usepackage{cvpr}
\usepackage{times}
\usepackage{epsfig}
\usepackage{graphicx}
\usepackage{amsmath}
\usepackage{amssymb}
\usepackage{float}
\usepackage{subcaption}
\usepackage{booktabs}
\usepackage{bbm}

\usepackage[pagebackref=true,breaklinks=true,letterpaper=true,colorlinks,bookmarks=false]{}

\cvprfinalcopy 

\def\cvprPaperID{****} 

\ifcvprfinal\fi
\begin{document}

\title{Visual Natural Language Query Auto-Completion\\ for Estimating Instance Probabilities}

\author{Samuel Sharpe\\
Columbia University\\
{\tt\small sbs2193@columbia.edu} 
\and
Jin Yan\\
Columbia University\\
{\tt\small jy2913@columbia.edu}
\and
Fan Wu\\
Columbia University\\
{\tt\small fw2321@columbia.edu }
\and
Iddo Drori\\
Columbia University\\
{\tt\small idrori@cs.columbia.edu}
}

\maketitle



\begin{abstract}
We present a new task of query auto-completion for estimating instance probabilities. We complete a user query prefix conditioned upon an image. Given the complete query, we fine tune a BERT embedding for estimating probabilities of a broad set of instances. The resulting instance probabilities are used for selection while being agnostic to the segmentation or attention mechanism. Our results demonstrate that auto-completion using both language and vision performs better than using only language, and that fine tuning a BERT embedding allows to efficiently rank instances in the image. In the spirit of reproducible research we make our data, models, and code available \footnote{\text{https://github.com/ssharpe42/VNLQAC}}.
\end{abstract}


\section{Introduction}
This work focuses on the problem of finding objects in an image based on natural language descriptions. Existing solutions take into account both the image and the query \cite{hu2016segmentation,Hu_2016_CVPR,shi2018key}. In our problem formulation, rather than having the entire text, we are given only a prefix of the text which requires completing the text based on a language model and the image, and finding a relevant object in the image. We decompose the problem into three components: (i) completing the query from text prefix and an image; (ii) estimating probabilities of objects based on the completed text, and (iii) segmenting and classifying all instances in the image. We combine, extend, and modify state of the art components: (i) we extend a FactorCell LSTM \cite{jaech2018personalized, jaech2018low} which conditionally completes text to complete a query from both a text prefix and an image; (ii) we fine tune a BERT embedding to compute instance probabilities from a complete sentence, and (iii) we use Mask-RCNN \cite{maskrcnn2017} for instance segmentation.

Recent natural language embeddings \cite{devlin2018bert} have been trained with the objectives of predicting masked words and determining whether sentences follow each other, and are efficiently used across a dozen of natural language processing tasks. Sequence models have been conditioned to complete text from a prefix and index \cite{jaech2018personalized}, however have not been extended to take into account an image. Deep neural networks have been trained to segment all instances in an image at very high quality \cite{maskrcnn2017,Hu_2018}.
We propose a novel method of natural language query auto-completion for estimating instance probabilities conditioned on the image and a user query prefix. Our system combines and modifies state of the art components used in query completion, language embedding, and masked instance segmentation. Estimating a broad set of instance probabilities enables selection which is agnostic to the segmentation procedure.
\section{Methods}
\begin{figure}
\centering
  \includegraphics[width=\linewidth]{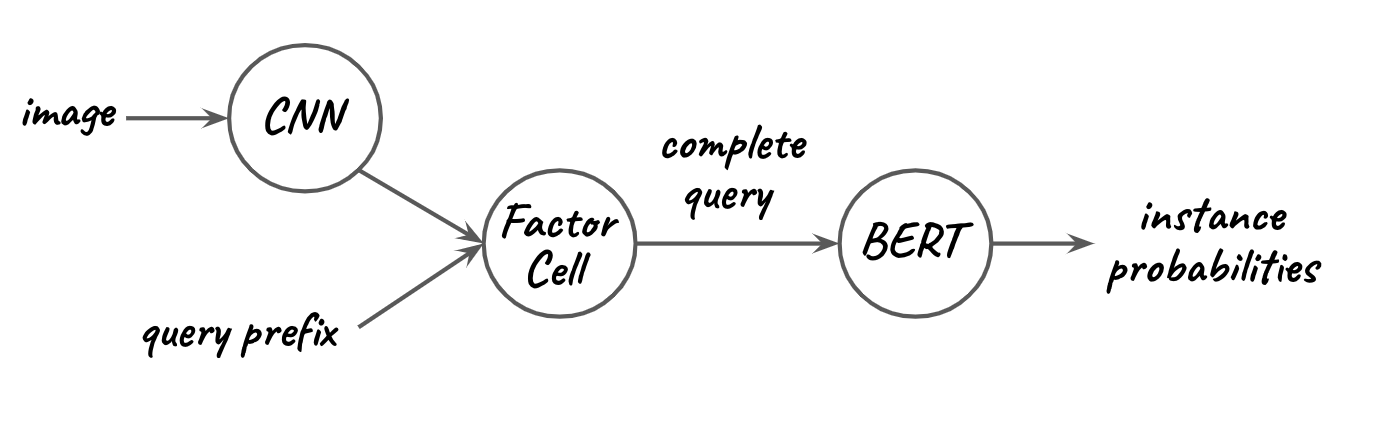}
  \caption{Architecture: Image features are extracted from a pretrained CNN along with the user query prefix are input to an extended FactorCell LSTM which outputs a completed query. The completed query is fed into a fine-tuned BERT embedding which outputs instance probabilities used for instance selection.}
  \label{fig:architecture}
\end{figure}

Figure \ref{fig:architecture} shows the architecture of our approach. First, we extract image features with a pre-trained CNN. We incorporate the image features into a modified FactorCell LSTM language model along with the user query prefix to complete the query. The completed query is then fed into a fine-tuned BERT embedding to estimate instance probabilities, which in turn are used for instance selection.
We denote a set of objects $o_k \in O$ where O is the entire set of recognizable object classes. The user inputs a prefix, $p$, an incomplete query on an image, $I$. Given $p$, we auto-complete the intended query $q$. We define the auto-completion query problem in equation \ref{eq:automax} as the maximization of the probability of a query conditioned on an image where $w_i \in A$ is the word in position $i$.
\vspace*{-.3cm}

\begin{equation}
    \label{eq:automax}
    \mathbf{q^*} = \underset{q}{\arg \max}{ P(q | p, I)} = \underset{\{w_1w_2\dots w_n\}}{\arg \max}{ P(w_1w_2\dots w_n| p, I)}
\end{equation}

We pose our instance probability estimation problem given an auto-completed query $\mathbf{q^*}$ as a multilabel problem where each class can independently exist. Let $O_{q*}$ be the set of instances referred to in $\mathbf{q^*}$. Given $\hat p_k$ is our estimate of $P(o_k \in O_{q*})$ and $y_k = \mathbbm{1}[o_k \in O_{q*}]$, the instance selection model minimizes the sigmoid cross-entropy loss function:
\begin{equation}
      \label{eq:select}
      \mathcal{L}_{selection} = -\sum_{k}{ y_k log(\hat p_k) + (1-y_k)log(1-\hat p_k) }.
\end{equation}


\subsection{Modifying FactorCell LSTM for Image Query Auto-Completion}
We utilize the FactorCell (FC) adaptation of an LSTM with coupled input and forget gates \cite{jaech2018low} to autocomplete queries. The FactorCell is an LSTM with a context-dependent weight matrix $\mathbf{W'} = \mathbf{W} + \mathbf{A}$ in place of $\mathbf{W}$. Given a character embedding $w_t \in \mathbb{R}^e$, a previous hidden state $h_{t-1} \in \mathbb{R}^h$ , the adaptation matrix, $\mathbf{A}$, is formed by taking the product\footnote{$\times_i$ represents the ith-mode tensor product. In other words, $\mathbf{Z_L}$ is reshaped to $\mathbb{R}^{m\times(r(e+h))}$ and $\mathbf{Z_R}$ is reshaped to $\mathbb{R}^{rh \times m}$} of the context, c, with two basis tensors $\mathbf{Z_L} \in \mathbb{R}^{m\times(e+h)\times r}$ and  $\mathbf{Z_R} \in \mathbb{R}^{r\times h \times m}$.

\begin{align}
\begin{split}
h_t & = \sigma([w_t, h_{t-1}]\mathbf{W'} + b) \\
\mathbf{A}&  = (c \times_1 \mathbf{Z_L}  )(\mathbf{Z_R} \times_3 c)
\end{split}
\end{align}

To adapt the FactorCell \cite{jaech2018low} for our purposes, we replace user embeddings with a low-dimensional image representation. Thus, we are able to modify each query completion to be personalized to a specific image representation. We extract features from an input image using a CNN pretrained on ImageNet, retraining only the last two fully connected layers. The image feature vector is fed into the FactorCell through the adaptation matrix. We perform beam search over the sequence of predicted characters to chose the optimal completion for the given prefix. 

\subsection{Fine Tuning BERT for Instance Probability Estimation}
We fine tune a pre-trained BERT embedding to perform transfer learning for our instance selection task. We use a 12-layer implementation which has been shown to generalize and perform well when fine-tuned for new tasks such as question answering, text classification, and named entity recognition. To apply the model to our task, we add an additional dense layer to the BERT architecture with 10\% dropout, mapping the last pooled layer to the object classes in our data. 

\subsection{Data and Training Details}

We use the Visual Genome (VG) \cite{krishnavisualgenome} and ReferIt \cite{KazemzadehOrdonezMattenBergEMNLP14} datasets which are suitable for our purposes. The VG data contains images, region descriptions, relationships, question-answers, attributes, and object instances. The region descriptions provide a replacement for queries since they mention various objects in different regions of each image. However, while some region descriptions are referring phrases, some are more similar to descriptions (see examples in Table \ref{table:region}). The large number of examples makes the Visual Genome dataset particularly useful for our task. The smaller ReferIt dataset consists of referring expressions attached to images which more closely resemble potential user queries of images. We train separate models using both datasets.

\begin{table}[H]
  \centering
 \begin{tabular}{lll}
    \toprule
    \cmidrule(r){1-2}
    Referring descriptions     &  Non-referring descriptions     \\
    \midrule
    guy sitting on the couch & couch is brown \\
    photos on white wall   &  small vehicle is van \\
    white keyboard on the desk  & mouse is in the charger \\
    \bottomrule
  \end{tabular}
  \caption{Example region descriptions from VG dataset.}
\label{table:region}
\end{table}

For training, we aggregated (query, image) pairs using the region descriptions from the VG dataset and referring expressions from the ReferIt dataset. Our VG training set consists of 85\% of the data: 16k images and 740k corresponding region descriptions. The Referit training data consists of 9k images and 54k referring expressions.

The query completion models are trained using a 128 dimensional image representation, a rank $r=64$ personalized matrix, 24 dimensional character embeddings, 512 dimensional LSTM hidden units, and a max length of 50 characters per query, with Adam at a 5e-4 learning rate, and a batch size of 32 for 80K iterations. The instance selection model is trained using (region description, object set) pairs from the VG dataset resulting in a training set of approximately 1.73M samples. The remaining 300K samples are split into validation and testing. Our training procedure for the instance selection model fine tunes all 12 layers of BERT with 32 sample batch sizes for 250K iterations, using Adam and performing learning rate warm-up for the first 10\% of iterations with a target 5e-5 learning rate. The entire training processes takes around a day on an NVIDIA Tesla P100 GPU.

\section{Results}
Figure 3 shows example results. We evaluate query completion by language perplexity and mean reciprocal rank (MRR) and evaluate instance selection by F1-score. We compare the perplexity on both sets of test queries using corresponding images vs. random noise as context. Table \ref{table:perplexity} shows perplexity on the VG and ReferIt test queries with both corresponding images and random noise. The VG and ReferIt datasets have character vocabulary sizes of 89 and 77 respectively.
\begin{table}[!htbp]
  \centering
  \begin{tabular}{cccc}
    \toprule
   Context Type & Visual Genome &  ReferIt \\
    \midrule
    Image & 2.38 & 2.66 \\
    Noise & 3.84 & 3.49 \\
    \bottomrule
  \end{tabular}
  \caption{Comparison of image query auto-completion perplexity using an image vs. noise, for both datasets. As expected, using the image results in a lower (better) perplexity.}
  \label{table:perplexity}
\end{table}
Given the matching index $t$ of the true query in the top 10 completions we compute the MRR as $\sum_{n}{\frac{1}{t}}$ where we replace the reciprocal rank with 0 if the true query does not appear in the top ten completions. We evaluate the VG and ReferIt test queries with varying prefix sizes and compare performance with the corresponding image and random noise as context. MRR is influenced by the length of the query, as longer queries are more difficult to match. Therefore, as expected we observe better performance on the ReferIt dataset for all prefix lengths. Finally, our instance selection achieves an F1-score of 0.7618 over all 2,909 instance classes.


\begin{figure}[!htbp]
\centering
  \includegraphics[width=\linewidth]{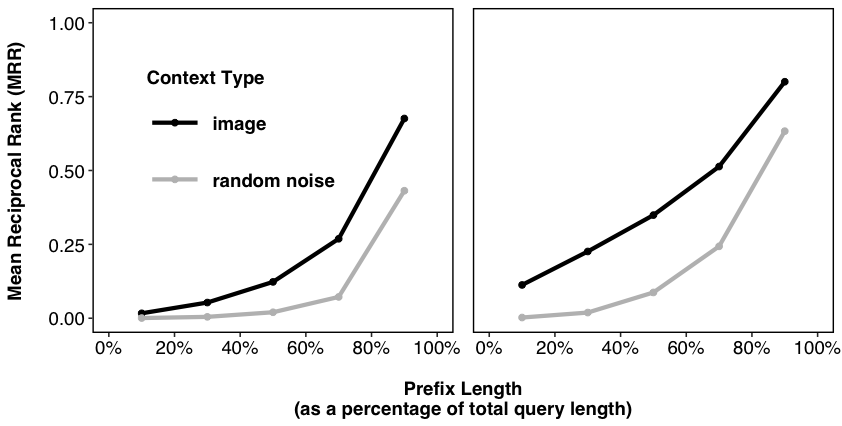}
    \caption{Comparison of image query auto-completion MRR results for VG (left) and ReferIt (right) using the image vs. noise. The horizontal axis denotes varying prefix lengths as a percentage of total query length and context. The MRR improves when increasing query prefix length, and is better when using the image.}
  \label{fig:mrr}
\end{figure}


\subsection{Conclusions}
Our results demonstrate that auto-completion based on both language and vision performs better than by using only language, and that fine tuning a BERT embedding allows to efficiently rank instances in the image. 
In future work we would like to extract referring expressions using simple grammatical rules to differentiate between referring and non-referring region descriptions. We would also like to combine the VG and ReferIt datasets to train a single model and scale up our datasets to improve query completions.

{\small
\bibliographystyle{plain}
\bibliography{bibliography.bib}
}

\begin{figure*}
\vspace{-20pt}
\centering
\begin{tabular}{cccc}
\vspace{5pt}
\subcaptionbox*{\bf{Query prefix: "wine b..."}}{\includegraphics[width = 1.5in]{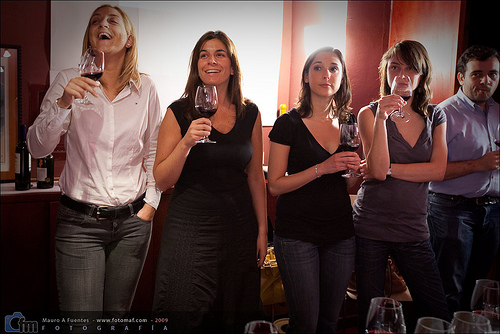}} &
\subcaptionbox*{\bf{Instance probabilities}}{\includegraphics[width = 1.5in]{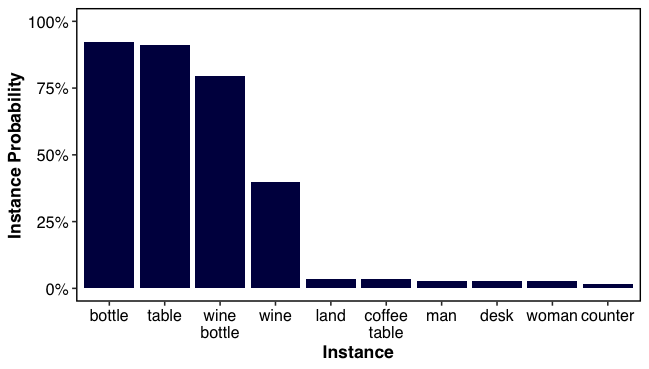}} &
\subcaptionbox*{\bf{Instance segmentation}}{\includegraphics[width = 1.5in]{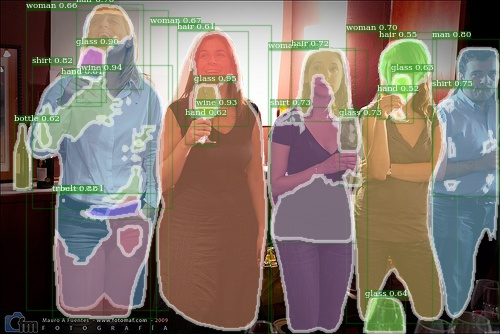}} &
\subcaptionbox*{\bf{Query completion: "wine bottle on table" and instance selection}}{\includegraphics[width = 1.5in]{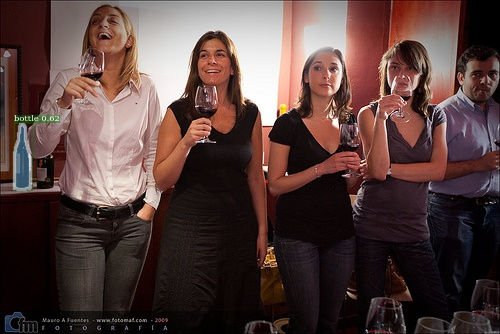}}\\
\vspace{5pt}
\subcaptionbox*{\bf{"a w..."}}{\includegraphics[width = 1.5in]{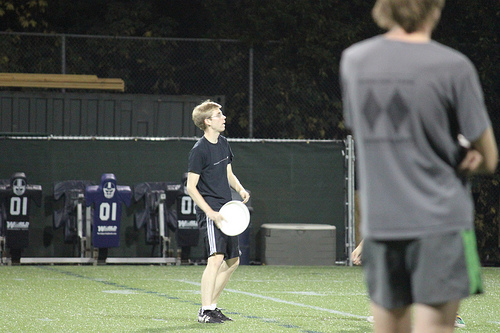}} &
\subcaptionbox*{}{\includegraphics[width = 1.5in]{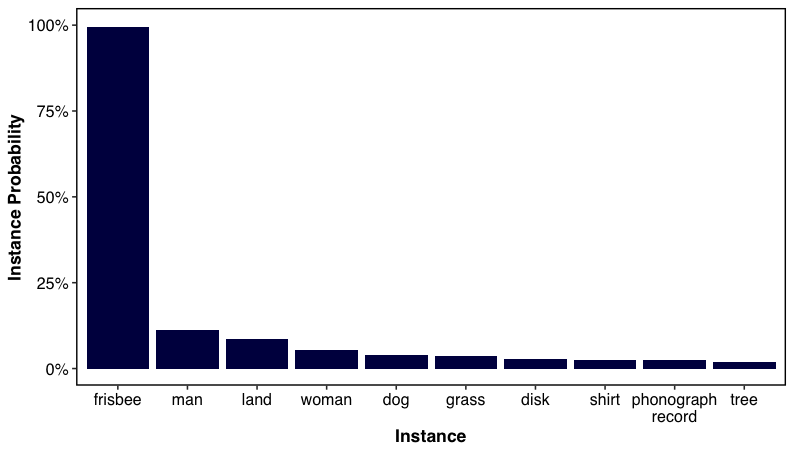}} &
\subcaptionbox*{}{\includegraphics[width = 1.5in]{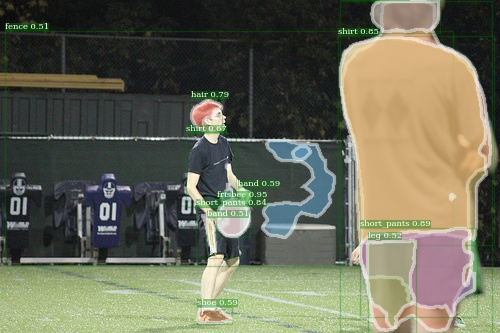}} &
\subcaptionbox*{\bf{"a white frisbee"}}{\includegraphics[width = 1.5in]{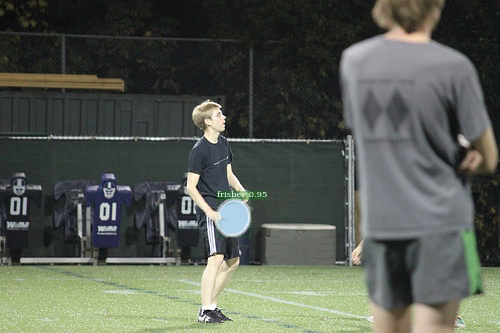}}\\
\vspace{5pt}
\subcaptionbox*{\bf{"brown h..."}}{\includegraphics[width = 1.5in]{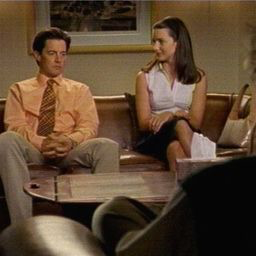}} &
\subcaptionbox*{}{\includegraphics[width = 1.5in]{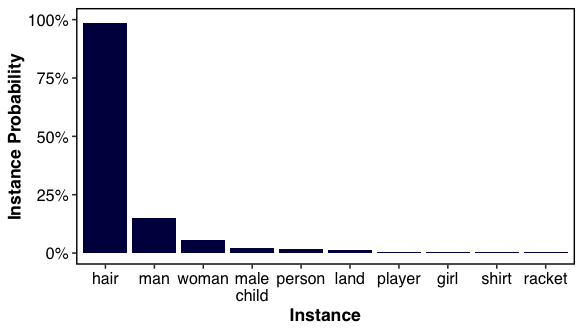}} &
\subcaptionbox*{}{\includegraphics[width = 1.5in]{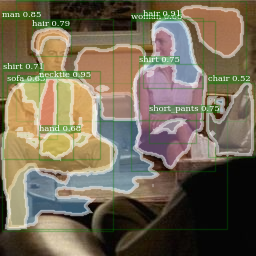}} &
\subcaptionbox*{\bf{"brown hair"}}{\includegraphics[width = 1.5in]{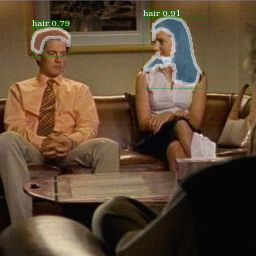}}\\
\vspace{5pt}
\subcaptionbox*{\bf{"he..."}}{\includegraphics[width = 1.5in]{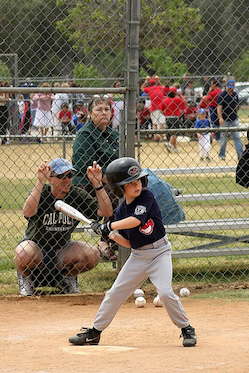}} &
\subcaptionbox*{}{\includegraphics[width = 1.5in]{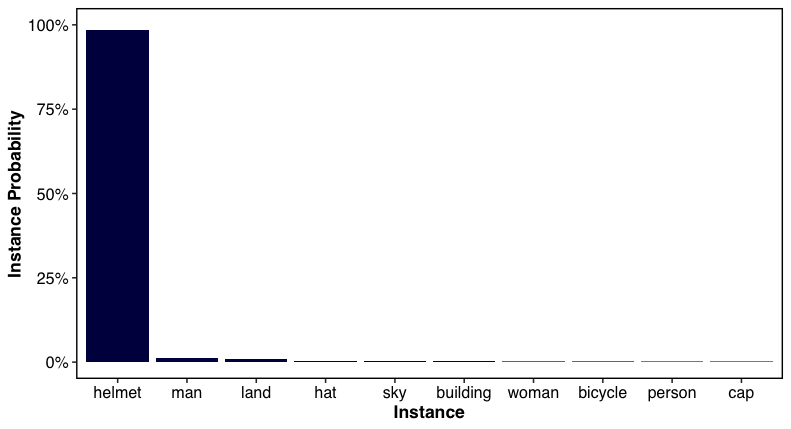}} &
\subcaptionbox*{}{\includegraphics[width = 1.5in]{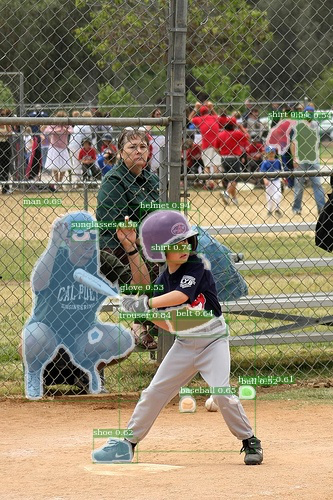}} &
\subcaptionbox*{\bf{"helmet is black"}}{\includegraphics[width = 1.5in]{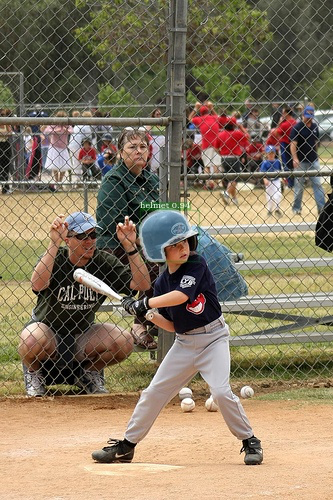}}\\
\vspace{5pt}
\subcaptionbox*{\bf{(a) Query prefix: "b..."}}{\includegraphics[width = 1.5in]{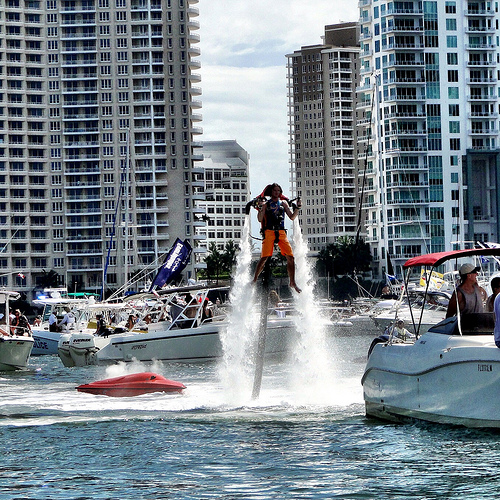}} &
\subcaptionbox*{\bf{(b) Instance probabilities}}{\includegraphics[width = 1.5in]{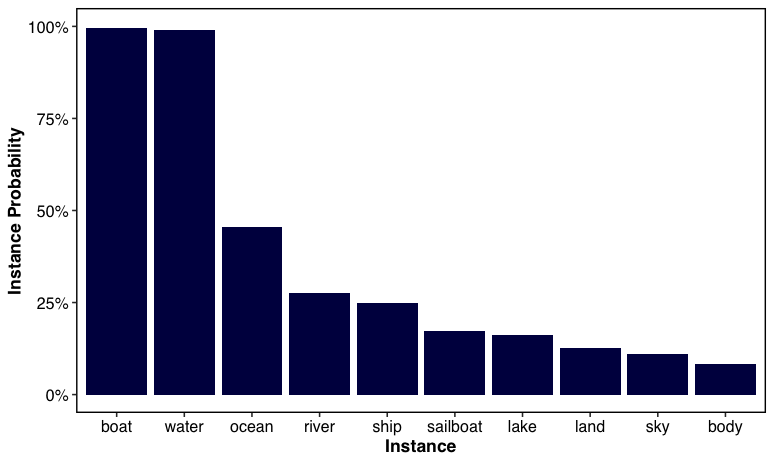}} &
\subcaptionbox*{\bf{(c) Instance segmentation}}{\includegraphics[width = 1.5in]{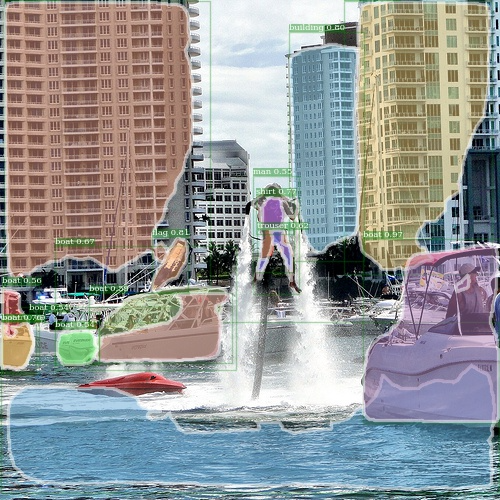}} &
\subcaptionbox*{\bf{(d) Query completion: "boats in water" and instance selection}}{\includegraphics[width = 1.5in]{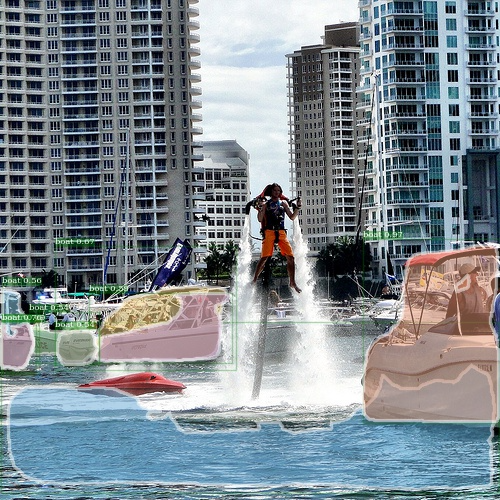}}\\
\end{tabular}
\caption{Example results: (a) input query prefix and image; (b) estimated instance probabilities; (c) instance segmentation; (d) resulting selected instances and auto-completed query conditioned on query prefix and image.}
\label{fig:examples}
\end{figure*}

\end{document}